\documentclass[journal]{IEEEtran}

\usepackage{cite}
\usepackage{mathtools}
\usepackage[table,x11names]{xcolor}
\usepackage{color, colortbl}
\definecolor{LRed}{rgb}{.8,1,.8}
\usepackage{graphicx}
\usepackage{epstopdf}
\usepackage{enumerate}
\graphicspath{ {./imgs/} }

\DeclarePairedDelimiter\floor{\lfloor}{\rfloor}
\ifCLASSINFOpdf

\else

\fi

\begin{document}
%
\title{A Neuromorphic Proto-Object Based Dynamic Visual Saliency Model with an FPGA Implementation}
%
%
%

\author{Jamal~Lottier~Molin,~\IEEEmembership{Member,~IEEE,}
		Chetan~Singh~Thakur,~\IEEEmembership{Senior Member,~IEEE,}
		Ernst~Niebur,
        Ralph~Etienne-Cummings,~\IEEEmembership{Fellow,~IEEE,}
\thanks{J. L. Molin is with the Machine Learning and Artificial Intelligence Lab, Riverside Research, 2640 Hibiscus Way, Beavercreek, OH 45431 USA e-mail: jmolin@riversideresearch.org.  C. S. Thakur is with the Department of Electronic Systems Engineering, Indian Institute of Science, Bangalore-560012 India e-mail: csthakur@iisc.ac.in.  E. Niebur is with the Department of Neuroscience, Johns Hopkins University, Baltimore, MD 21218 USA e-mail: niebur@jhu.edu.  R. Etienne-Cummings is with the Department of Electrical and Computer Engineering, Johns Hopkins University, Baltimore,
MD, 21218 USA e-mail: retienne@jhu.edu}
\thanks{Manuscript received January 1, 2020; revised Jan 2, 2020.}}

\maketitle

\begin{abstract}
The ability to attend to salient regions of a visual scene is an innate and necessary preprocessing step for both biological and engineered systems performing high-level visual tasks (e.g. object detection, tracking, and classification).  Computational efficiency, in regard to processing bandwidth and speed, is improved by only devoting computational resources to salient regions of the visual stimuli.  In this paper, we first present a neuromorphic, bottom-up, dynamic visual saliency model based on the notion of proto-objects.  This is achieved by incorporating the temporal characteristics of the visual stimulus into the model, similarly to the manner in which early stages of the human visual system extracts temporal information.  This neuromorphic model outperforms state-of-the-art dynamic visual saliency models in predicting human eye fixations on a commonly-used video dataset with associated eye tracking data.  Secondly, for this model to have practical applications, it must be capable of performing its computations in real-time under low-power, small-size, and lightweight constraints.  To address this, we introduce a Field-Programmable Gate Array implementation of the model on an Opal Kelly 7350 Kintex-7 board.  This novel hardware implementation allows for processing of up to 23.35 frames per second running on a 100 MHz clock -- better than 26$\times$ speedup from the software implementation.
\end{abstract}


%
\IEEEpeerreviewmaketitle

\section{Introduction}
\IEEEPARstart{I}n the field of neuromorphic engineering, we seek to design systems which mimic the mechanisms of the human brain.  The human visual system (HVS) is capable of efficiently
performing complex visual tasks in real-time under low size, weight, and power (SWaP) constraints.  In this work, we have designed our neuromorphic saliency model based on the neurophysiological properties observed in the HVS.  By doing so, we further bridge the gap between engineered systems and the human brain.

In th HVS, each optic nerve
receives input from retinal ganglion cells transmitting neural
information to the brain in the form of spikes, otherwise referred to
as action potentials \cite{koch2004efficiency}.  The rate at which
these cells transmit neural information is equivalent to the brain
receiving $\sim100$ Mbps of spatial and temporal visual input per
optic nerve \cite{strong1998entropy}.  Processing this overwhelming
amount of data in parallel, and in real-time, is impossible for any
human brain.  To overcome this complexity, the HVS instead utilizes selective
attention and attends to only regions of the visual stimuli deemed
interesting, or salient.  It is these salient regions that are then
forwarded to the succeeding stages of processing.  This idea is known
as visual saliency.  There are two components of visual saliency, bottom-up and top-down.  Bottom-up saliency is a function of only the inherent properties of
the visual stimulus itself.  Top-down saliency is a function of the viewer's biases based on their internal state and goals.  In this work, we address bottom-up
saliency.  The proper computation of visual saliency also serves as an
aid in the field of computer vision.  It is essential for any system
performing higher-level visual tasks including, but not limited to,
navigation and localization, object tracking and classifation,
image/video compression, surveillance and security, and action
recognition.  The ability to determine salient, interesting regions of
the visual scene increases computational efficiency by minimizing
throughput, reducing data dimensionality, and increasing overall
processing speed \cite{park2013saliency, wang2011saliency, riesenhuber1999hierarchical}.  Finally, modeling dynamic visual
saliency in a biologically-plausible manner is ideal for the most
efficient visual saliency computational system, especially for an
engineered system which seeks to emulate biology.  As in biology, it
is important that motion is considered when computing saliency, rather
than computation on only static stimuli.  In this model, we extend
prior preliminary work \cite{molin2013proto,molin2015motion}, and we
take a biologically-plausible, bottom-up approach for computing
dynamic visual saliency which considers motion exhibited within the
scene.

There exists psychophysical and neurophysiological evidence
of the existence of a retintopic, saliency map computed within
the HVS \cite{duncan1984selective, gottlieb1998representation, robinson1992pulvinar, qiu2007figure}.  Inspired by the Feature Integration Theory of Attention \cite{treisman1980feature} and Koch and
Ullman \cite{koch1987shifts}'s work on selective attention, Itti, Koch, and Niebur
designed a biologically-plausible computational model of feature-based visual attention \cite{itti1998model}.  This model has significantly influenced the
field of visual attention as many succeeding visual saliency models
are derivations from this Itti et al. model.  On the other hand, saliency models exist which are supported by Gestalt psychology, based on the idea the whole is perceived before the parts (i.e. features) \cite{sun2008computer, walther2006modeling, russell2014model, yanulevskaya2013proto}.
These models are referred to as object-based saliency models to emphasize the idea that attention does not depend solely on image features, but
rather on the structural organization of the scene into perceptual
objects.  This approach to computing visual saliency is backed by
neurophysiological and psychophysical evidence demonstrating objects
are in fact perceived prior to features
\cite{cave1999visuospatial,duncan1984selective,qiu2007figure}.  One
hypothesis explaining object-based attention is the coherence theory
suggested by Rensink \cite{rensink2000dynamic}.  The coherence theory
states low-level proto-objects exist which are formed
rapidly and in parallel across the visual field.  These proto-objects
are pre-attentive structures with limited spatial and temporal
coherence.  Focused attention is required to stabilize a small number
of proto-objects, and therefore, generating the perception of an object
with a much higher degree of coherence over space and time.  Once
attention is released, the object dissolves back to its dynamic
proto-object state \cite{rensink2000dynamic}.  Proto-objects can be better understood as the highest-level output of low-level vision and the lowest-level
operand on which higher-level process can act, including visual
attention. 

\begin{figure}[ht]
\centering
\includegraphics[width=0.45\textwidth]{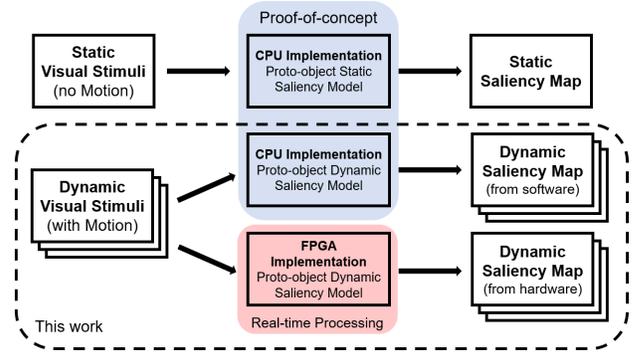}
\caption{The dotted line signifies our contributions -- the proto-object based dynamic saliency model discussed in this paper.  Contents outside the box represent previous work by Russell et al. \cite{russell2014model}.  The blue transparent box signifies the CPU/software proto-object based saliency models (both static and dynamic).  The red transparent box signifies the FPGA implementation (for real-time, low SWaP) of our dynamic visual saliency model, suitable for real-world applications.}
\label{fig:this_work}
\end{figure}

To optimize the computational efficiency and accuracy of this dynamic visual saliency model, it is imperative to keep the model biologically-plausible with respect to its computations.  To achieve this, we extend the proto-object based saliency model by Russel et al. \cite{russell2014model} -- a bottom-up, feed-forward computational model of visual saliency that computes saliency as a function of figure-ground relationships attained via the notion of proto-objects.  Although biologically-plausible and capable of predicting human eye fixations on static scenes better than other state-of-the-art (SOTA) saliency models, the Russel et al. model did not take into account the temporal characteristics within visual stimuli.  Taking this into account is critical given motion is the most significant contributor when computing visual saliency \cite{mahapatra2008motion}.  The major contributions of this work are as follows (see Fig.~\ref{fig:this_work}):

\begin{enumerate}
\item We present a neuromorphic, dynamic visual saliency model, which considers motion in
  the visual scene.  The idea of motion is integrated into the
  proto-object based visual saliency model in a biologically-plausible
  manner that is sufficient for computing bottom-up visual saliency on
  videos (i.e. dynamic scenes).  This model is based solely on neurophysiological properties of the HVS, thus parameters are fixed and no training is required.  This is an immense advantage over conventional machine learning based models, which suffer in the case of data unavailability.
\item We further introduce a novel hardware implementation of this
  dynamic proto-object based visual saliency model on an FPGA (Field
  Programmable Gate Array) making possible the real-time and low SWaP processing that we seek.  The system is capable of predicting human eye fixations on static visual stimuli and dynamic visual stimuli, making it sufficient for real-world applications.
\end{enumerate}

\section{Related Work}
\label{sec:related_work}
\subsection{Current Dynamic Visual Saliency Models}
Many current models of visual saliency, both object-based and
feature-based, compute saliency only on static visual stimuli and do
not consider motion that may exist within the visual scene.  To
validate these saliency models, datasets of static images with
corresponding human eye fixation data are used to quantify the extent
to which the saliency model predicts eye fixations.  However, the
world is dynamic and constantly changing.  Motion is a naturally
occurring phenomenon that plays an important role in both human and computer visual
processing, and specifically, in visual attention.  For human observers, it has been shown that given a dynamic visual stimulus, motion plays a more
significant role in visual saliency than other low-level features
\cite{mahapatra2008motion}.  Thus, it is important to consider the
temporal dynamics of the visual stimuli when computing visual
saliency.  More recently, saliency models have been implemented that
do consider motion when computing a dynamic saliency map. 

Rosenholtz \cite{rosenholtz1999simple} introduced a dynamic saliency model which interprets saliency as an outlier to a statistical distribution of motion features.  Similarly, Gao et al. \cite{gao2008plausibility} developed a model which considers motion using biologically-plausible spatio-temporal Gabor filters and computes the Kullback Leibler divergence between distributions of pixel feature responses from the pixel’s local region.  Seo et al. \cite{seo2009static} proposed a self-resemblance method for computing saliency using statistics to measure likelihood of saliency at a given pixel relative to its local neighborhood.  Itti and Baldi \cite{itti2002real,itti2005quantifying} introduced a saliency model which considers motion as an additional feature using Bayesian surprise.  This model is also based on statistics and uses Bayes' theorem to statistically compute how much a new
observation differs from its prior.  Zhang et al. \cite{zhang2009sunday} computed saliency on dynamic scenes by computing dynamic saliency based on statistics and requires learning the probability distribution for each spatio-temporal feature.

Itti et al. \cite{itti2005quantifying} extended their original model to
include two additional feature channels: flicker (on-set/off-set) and motion
channels.  Harel et al. \cite{harel2007graph} reformulated this Itti et al. method from a graph-based perspective.  In its original state, this model did not consider motion.  However, its software implementation included the option for a motion channel \cite{harel2006saliency}.  Nonetheless, this graph-based approach is less biologically-plausible in its computation.  Other feature-based dynamic saliency models which consider motion include \cite{marat2009modelling,guo2010novel, leboran2016dynamic, muddamsetty2018salient}.

Some recent feature-based models are learning-based in order to compute dynamic saliency \cite{bak2017spatio,liu2011learning}.  However, these learning methods may not be biologically-plausible in their computations.

In this work, we present a neuromorphic, bottom-up, object-based dynamic visual
saliency model based on the notion of proto-objects using spatial and
temporal filters.  There is no learning required for this model as it
is based on neurophysiological evidence.  This model is
biologically-plausible, using temporal filters similar to the receptive fields of simple cells observed in visual cortex V1 in the parvoceullar and magnocelluar pathways.  Furthermore, this model
builds on the proto-object saliency model which supports the idea that
objects are perceived prior to features.  Consequently, saliency is
computed as a function of dynamic proto-objects existing within the
visual field, opposed to only from features.

\subsection{Current FPGA-based, Real-time Saliency Models}
Considering the computational complexity of this model, we accelerate the computation of the dynamic visual saliency map using a novel FPGA implementation.  This allows for real-time processing of a reliable, biologically-plausible dynamic visual saliency model that is capable of predicting human eye fixations better than other SOTA models.  Such real-time processing allows for integration with other visual processors that require such rapid higher-level processing including object recognition and detection.

All of the visual saliency models previously discussed were implemented in software and run on CPUs for proof-of-concept.  Over the past decade, there has been increasing interest in implementing visual saliency models on FPGAs for real-time, low-power processing.  Bouganis et al. \cite{bouganis2006fpga} accelerated a saliency model proposed by Li et al. \cite{li1999visual} which operated on the
gray-scale of a single image only and utilized neuron models tuned to
specific orientation and spatial locations.  The differential
equations used to model these neurons for computing saliency are computationally-demanding and therefore, the array of neurons with their associated dynamics was implemented on FPGA.  Using the parallel architecture of the FPGA demonstrated a speedup of more than $10\times$.  

Kestur et al. \cite{kestur2012emulating} utilized FPGA to implement a
library for saliency computation based on the Itti et al. (1998) model
\cite{itti1998model}.  This FPGA-based accelerator is called Streaming
Hardware Accelerator with Run-time Configurability (SHARC) and showed
$5\times$ speedup to CPU-based version of the saliency model on $256
\times 256$ images.  Akselrod et al. \cite{akselrod2011hardware} utilized their NeuFlow platform of implementing a simplified Itti et al. saliency model showing a $4 \times$ speedup on $480 \times 480$ images in comparison to CPU
implementation.  Motion was also incorporated into the model.  Kim et
al. \cite{kim2011implementation} also implemented the model on FPGA,
simplifying the normalization operation for FPGA.  Their
implementation interfaces with a silicon retina chip and extracts
various features on $128 \times 128$ images.  They were able to show a
speedup of more than $2.5\times$ and power reduction of more than
$32\times$ by using the FPGA implementation.  Moradhasel et
al. \cite{moradhasel2013fast} designed an FPGA based saliency model
and showed computation speeds of 50 million pixels per second.
Similarly to the Akselrod et al. \cite{akselrod2011hardware}
implementation, they also considered motion in this model.  It showed
a $2\times$ speedup over state-of-the-art models at that time.  Most
recently, Barranco et al. \cite{barranco2014real} developed a
simplified, yet more complete FPGA implementation of the saliency
model incorporating motion as well as winner-take-all and inhibition
of return.  It also has a top-down component which modulates the final
saliency map as a function of optical flow and depth.  This model
outperformed all previous models with respect to speed as it computed
saliency maps at 180 fps for $640 \times 480$ resolution.  Finally,
other saliency models have been implemented on FPGA including the work
of Bae et al. \cite{bae2011fpga} where the AIM (Attention based on
Information Maximization) algorithm by Bruce et
al. \cite{bruce2007attention} was implemented on an FPGA platform for
real-time processing capable of 4 million pixels per second.

The FPGA implementations of saliency models discussed demonstrate the
advantages that FPGA implementations have over CPU implementations
with regard to processing speed and SWaP.  However, all of these
models are purely feature-based.  The FPGA implementation presented here is, to our knowledge, the first proto-object based model implemented on non-CPU hardware.

\section{Original Proto-Object Based Saliency Model for Static Images}
\label{sec:russell_model}
Our model is inspired by the original model of proto-object based saliency by Russell et al. \cite{russell2014model} for static images.  Therefore, it is important to discuss this original model prior to discussing our saliency model for dynamic visual stimuli (i.e. video).  The Russel et al. model is an object-based, bottom-up, feed-forward model of visual saliency.  It is based on the notion of proto-objects which may exist within the visual field.  The model outperformed other state-of-the-art models \cite{harel2007graph,itti1998model} of visual saliency on predicting human eye fixations on a dataset of static images of natural scenes.  The model works as follows: It receives an RGB image (resolution of $640 \times 480$) and decomposes this image into three feature channels: intensity, color, and orientation.  Within each of these feature channels are sub-channels.  The intensity channel has one sub-channel.  The color channel has four sub-channels: red-green opponency, green-red opponency, blue-yellow opponency, and yellow-blue opponency.  The orientation channel also has four sub-channels (four orientations): $0^\circ$, $45^\circ$, $90^\circ$, and $135^\circ$.  This results in a total of nine feature channels.  Once the original color image is decomposed into these nine channels, within each channel, the feature map is successively down-sampled in steps of $\sqrt{2}$ to form an image pyramid spanning five octaves.  Proto-object activity is then computed within each channel and at each level of the pyramid, independently.  Proto-object activity gives rise to saliency with respect to figure-ground relationship within the visual scene.  The proto-objects are computed using a grouping mechanism consisting of edge and center-surround operators working together to compute border ownership activity.  Neurons encoding border ownership (one-sided assignment of a border to a region perceived as a figure) have been discovered in early stages of visual processing, predominantly in visual cortex V2, by Zhou et al. \cite{zhou2000coding}.  This border ownership activity is integrated in a circular fashion to reveal grouping activity.  More details on this grouping mechanism can be found in \cite{russell2014model}.  A normalization operation, $N_1$, is then applied to each grouping activity map to enhance maps with single proto-objects and suppress maps with multiple proto-objects.  This normalization operator, $N_1$, works as follows:
\begin{enumerate}
\item The maximum, $m$, of the map being normalized is determined.
\item The average of the other local maxima, $\bar{m}$, is determined.
\item Finally, there is a global, element-wise multiplication of the map by $(m-\bar{m})^2$.
\end{enumerate}
This normalization, $N_1$, is a function of the grouping activity such that grouping activity with few proto-objects is promoted while grouping activity of maps with multiple proto-objects is suppressed.  Following this normalization, a similar computation to that in the Itti et al. (1998) model \cite{itti1998model} is performed.  The image pyramids within each channel are collapsed by scaling each level to a common level and summing.  This results in a single conspicuity map within each channel.  These nine conspicuity maps are then normalized using a second (but similar) normalization operator, $N_2$.  The normalization operator, $N_2$, works as follows:
\begin{enumerate}
\item The map is normalized to the range $[0,...,M]$.
\item The maximum, $m$, of the map being normalized is determined.
\item The average of the other local maxima, $\bar{m}$, is determined.
\item Finally, there is a global, element-wise multiplication of the map by $(m-\bar{m})^2$.
\end{enumerate}
The only difference in this normalization, $N_2$, is the additional first step of normalizing each map to a common range $[0,...,M]$.  This step is necessary for allowing invariance to modality (feature).  This globally enhances conspicuity maps with few strong peak responses and globally suppresses maps with many comparable peak responses.  Finally, these normalized conspicuity maps are linearly summed to form the final saliency map.

\section{Our Neuromorphic Model}
\label{sec:podvs_model}
We name our neuromorphic model of dynamic visual saliency PODVS (Proto-object Based Dynamic Visual Saliency).  It utilizes the idea of separable space-time filters for incorporating motion.  This biologically-plausible model is based on the idea that simple cells in the magnocelluar and parvocellular pathways act as spatio-temporal filters.  They not only extract spatial information preattentively, but also temporal information.  This model utilizes the ideas of the original proto-object based visual saliency model by Russel et al. \cite{russell2014model}, however, now extracts both temporal and spatial information for computing saliency on dynamic visual stimuli.  This allows for computing saliency in videos by considering motion that may exist within the scene.

In the following sections, we will discuss our model, starting with how motion is computed in the HVS, followed by how we use such biological motion computation in the our proto-object based model.  After discussing our model, we will introduce our novel FPGA implementation of this model for real-time, low SWaP processing.

\subsection{Motion Processing in the Human Visual System}
We seek to represent motion in our model in a biologically-plausible manner.  This model is purely bottom-up and feed-forward.  Henceforth, we focus on how motion is computed at early stages of visual processing, and further, preattentive visual processing.  Neurophysiological research has shown that motion extraction occurs along the dorsal pathway beginning in V1 and proceeds to middle temporal area (MT) and then continues to the medial superior temporal area (MST) \cite{de2000spatial}.  Motion extraction in V1 can be represented by local spatio-temporal filters and shows preference to spatial frequency, spatial phase, spatial orientation, and direction of motion.  Later stages of motion are responsible for computing velocity and optical flow.  This motion processing requires attention \cite{cavanagh1992attention}. However, we are interested in representing preattentive motion, and therefore, consider motion extraction in V1 only.

It should be noted we are concerned with the receptive field of non-direction selective simple cells.  As previously noted, there are two pathways within the visual system: the magnocelluar and parvocelluar pathways \cite{de2000spatial}.  Each of these pathways has a select population of retinal cells, which project to the LGN (Lateral Geniculate Nucleus), and further to primary visual cortical cells.  Strongly phasic simple cells exist within the magnocellular pathway, they have high temporal resolution, high contrast sensitivity and low color sensitivity.  Cells in the parvocelluar pathway are weakly phasic and have low contrast sensitivity and high color sensitivity but low temporal resolution.  Strongly phasic cells typically have a strong excitatory phase followed by a strong inhibitory phase.  Weakly phasic cells typically have a less pronounced excitatory phase followed by a weak inhibitory phase, resulting in a weaker response to motion.  Approximately 20-25 percent of the population of non-direction selective simple cells in V1 are strongly phasic.  The remaining are weakly phasic.  The temporal filters used in the models to be discussed are modeled to fit the receptive fields of these strongly and weakly phasic cells found in the primary visual cortex \cite{hawken1996temporal,de2000spatial}.

\begin{figure}[!htbp]
\centering
\includegraphics[width=0.45\textwidth]{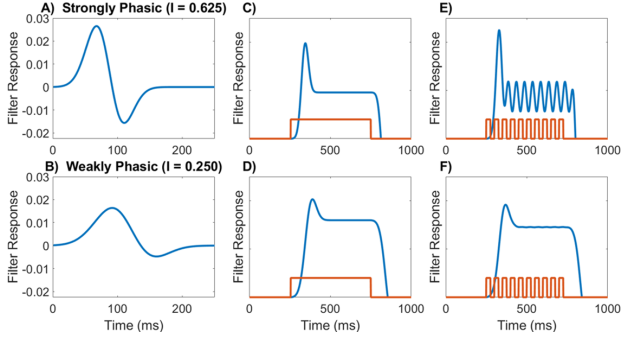}
\caption{Plots A and B show the temporal profile of a strongly phasic
  and weakly phasic filter, respectively.  Plots C and D show the
  filter response to an abrupt then constant stimulus.  Plots E and F show the filter response to flicker motion (continuous onset/offset change).  Strongly phasic filters are more sensitive to temporal change.  $I$ is the ratio of the peak positive to the peak negative amplitude of the filter, hence, representing the degree to which the filter is strongly phasic.  Image derived from \cite{parkhurst2002selective}.}
\label{fig:temporalphasicmodel}
\end{figure}

\subsection{Biologically-plausible Temporal Filters}
The work of Parkhurst \cite{parkhurst2002selective} and De Valois et al. \cite{de2000spatial} was used to model the transfer function for a biologically-plausible temporal filter modeling the temporal receptive field of strongly phasic and weakly phasic, non-direction selective simple cells in V1.  The approximation of the transfer function of the V1 simple cell temporal receptive field can be seen in Equation \ref{eq:temporalfilter}.

\begin{equation}
\label{eq:temporalfilter}
r(t)=\alpha(t-\tau-\delta)e^{\beta(t-\tau)^2}
\end{equation}

\begin{itemize}
\item $\alpha$ - response amplitude parameter
\item $\beta$ - response amplitude parameter
\item $\tau$ - time shift parameter
\item $\delta$ - determines degree to which weakly or strongly phasic in time
\end{itemize}

These parameters were fit to model the temporal response profile of strongly phasic and weakly phasic cells in V1 from neurophysiological recordings \cite{parkhurst2002selective}.  These parameter values can be seen in Table~\ref{tab:temporalparams}.

\begin{table}[!htbp]
\caption{Parameters for Strongly/Weakly Phasic V1 Simple Cell Temporal Response}
\label{tab:temporalparams}
\centering
\begin{tabular}{|c||c|c|c|c|}
\hline
\textbf{Type}		    & \textbf{$\alpha$} & \textbf{$\beta$} & \textbf{$\tau$} & \textbf{$\delta$} \\
\hline
\hline
Strongly Phasic & -0.00161 & -0.00111 & 86.2  & 5.6\\
\hline
Weakly Phasic & -0.000487 & -0.000466 & 116  & 20\\
\hline
\end{tabular}
\end{table}

These temporal profiles for strongly and weakly phasic simple cell
receptive fields are applied in our dynamic model of proto-object
based visual saliency.  The methodology in regards to how they are
applied will be discussed in Section~\ref{sec:featureextraction}.  A
visual representation of the filters can be seen in
Fig.~\ref{fig:temporalphasicmodel}.  The strongly phasic temporal
filter in Fig.~\ref{fig:temporalphasicmodel}A, has a strong
positive/excitatory lobe and a strong negative/inhibitory lobe, hence,
is more sensitive to motion.  The weakly phasic temporal filter in
Fig.~\ref{fig:temporalphasicmodel}B, has a positive/excitatory lobe
and weak negative/inhibitory lobe, and, hence, is less sensitive to motion.  While the $y$-axis is the filter coefficient, the $x$-axis is time (in the past) assuming a rate of 24 frames per second for the incoming input image sequence.

Fig.~\ref{fig:temporalphasicmodel} shows the response to various types of stimuli.  The value of $I$ is the ratio of the peak positive amplitude to the peak negative amplitude of the temporal filter, representing the degree to which the filter is strongly phasic.  Hence, the strongly phasic filter has a larger value for $I$ ($I=0.625$) and the weakly phasic filter has a smaller value for $I$ ($I=0.250$).  Fig.~\ref{fig:temporalphasicmodel}A and Fig.~\ref{fig:temporalphasicmodel}B are the strongly and phasic filters, respectively. Fig.~\ref{fig:temporalphasicmodel}C and Fig.~\ref{fig:temporalphasicmodel}D are their responses to an abrupt onset then constant stimulus.  Fig.~\ref{fig:temporalphasicmodel}E and Fig.~\ref{fig:temporalphasicmodel}F are their responses to flicker (onset/offset) motion.  The strongly phasic filter clearly has a higher response to temporal changes in the stimuli.

Because of the finite duration of the filter functions, frames more than 250 ms in the past do not contribute to  saliency. Components of the visual stimulus that do not change over an extended amount of time generate a lower temporal response. Relatedly,  these temporal filters have a stronger response to onset and offset of objects within the scene.  These temporal dynamics are similar to those seen in the model of visual saliency depicted in \cite{zhang2009sunday}, however, their model uses learning.  In the following sections we will discuss how these temporal filters are utilized within our model, specifically, how these temporal filters are applied within each feature channel for computing proto-objects.

\begin{figure*}[!htbp]
\centering
\includegraphics[width=.75\textwidth]{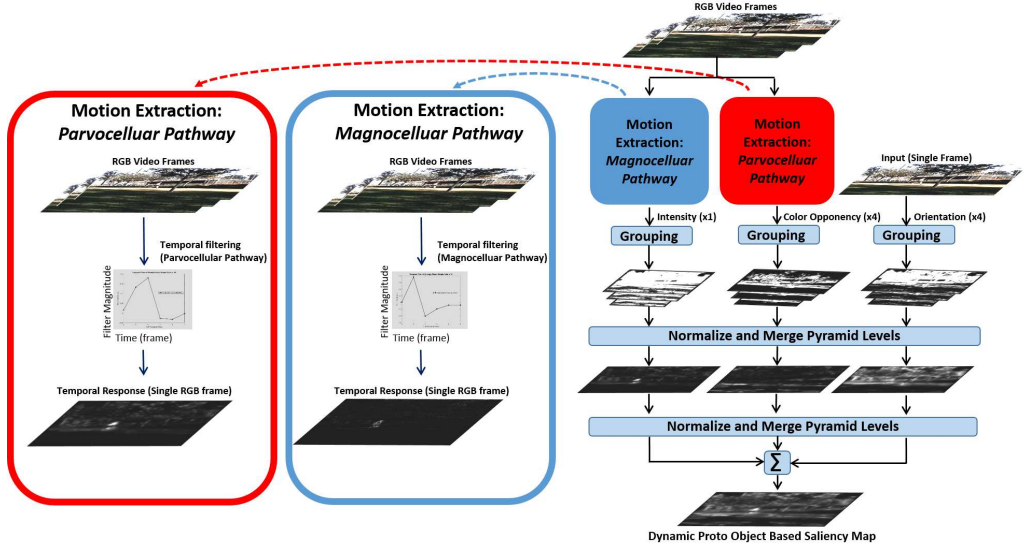}
\caption{PODVS - Proto-object based Dynamic Visual Saliency.  This model utilizes spatial and temporal information of the dynamic scene as input to the model.  The model receives RGB video frames as input.  For the intensity channel, motion is extracted using strongly phasic temporal filters (magnocelluar pathway).  The output of this response serves as input into the grouping stage for computing dynamic, pre-attentive proto-objects in the intensity channel.  In the color channels, weakly phasic temporal filters (parvocelluar pathway) are used, which are less sensitive to motion and retain more static regions of the scene.  The output of this spatio-temporal response serves as input to the grouping stage within the color channel.  No motion is extracted within the orientation channel, therefore, static information is preserved.  Details of this model can be found in Section \ref{sec:podvs_model}.}
\label{fig:dpovsst}
\end{figure*}

\subsection{Spatio-temporal Feature Channel Extraction}
\label{sec:featureextraction}
The model receives dynamic visual stimuli (i.e. color video) as input.  This input can be realized as a sequence of images (RGB video frames).  We assume a resolution of $640 \times 480$ and video frame rate of 24 frames per second.  For each frame, a new saliency map is computed in terms of dynamic proto-objects from spatio-temporal responses within different feature channels: intensity, color, and orientation.

\subsubsection{Intensity Channel}
Given that simple cells in the magnocelluar pathway have high contrast sensitivity, low color sensitivity, and high sensitivity to motion \cite{de2000spatial}, we apply the strongly phasic temporal filter in the intensity channel.  To extract the
intensity of the current frame, the average of the red (r), green (g),
and blue (b) channel is computed (see Equation \ref{eq:intensityextraction}).  The temporal filter is applied on the current frame and
previous frames of the intensity-valued frames.  The convolution is
applied temporally across the video frames.  The total number of
frames in which the convolution is applied is dependent on the frame
rate of the videos.  In our case, our chosen frame rate of 24 Hz
results in a filter convolution over the current frame and five
previous frames.  The representation of this discrete convolution can
be seen in Equation \ref{eq:discretetemporalfilter}. 

\begin{equation}
\label{eq:discretetemporalfilter}
M_S[n] = (F*R_S)[n] = \sum_{t=0}^{T}\sum_{r=1}^{N_r}\sum_{c=1}^{N_c}F_{r,c}[n-t] \times R_S[t]
\end{equation}

$M_S[n]$ is the strongly phasic temporal output at frame $n$.
$F_{r,c}[n]$ is the pixel intensity of the original grayscaled
(intensity version) video at row $r$ and column $c$ at frame $n$.
$R_S[t]$ is the discretized representation of the filter $r(t)$ in
Equation \ref{eq:temporalfilter} using the strongly phasic parameters
in Table \ref{tab:temporalparams}.  Finally, $F[n-t]$ represents the
frame at $t$ frames in the past.  $T$ is the total number of frames in
the past over which to perform the convolution. In our case, $T = 6$.
$N_r$ and $N_c$ are the number of rows and columns, respectively, in
each frame.  The output, $M_S[n]$ (strongly phasic output) is the
input to the grouping stage within the intensity channel. $M_S[n]$ has dimensions $N_r \times N_c$.

\subsubsection{Color Channel}
Given that simple cells in the parvocelluar pathway have low
contrast sensitivity, high color sensitivity, and are less sensitive to motion, the weakly
phasic temporal filter is applied within this channel.  The same
convolution is performed on the video sequence for the red, green, and
blue channels.  However, in this case, the discretized weakly phasic
filter is applied, and henceforth, $R_W[t]$ is modeled by Equation
\ref{eq:temporalfilter} using the weakly phasic parameters in Table
\ref{tab:temporalparams}.  The weakly phasic motion response can be
seen in Equation \ref{eq:discretetemporalfilter_weak}.  The only
difference between Equation \ref{eq:discretetemporalfilter} and
Equation \ref{eq:discretetemporalfilter_weak} is the temporal filter
used. 

\begin{equation}
\label{eq:discretetemporalfilter_weak}
M_W[n] = (F*R_W)[n] = \sum_{t=0}^{T}\sum_{r=1}^{N_r}\sum_{c=1}^{N_c}F_{r,c}[n-t] \times R_W[t]
\end{equation}

The output, $M_W[n]$ (weakly phasic output), is the input to that of the color channel.  $M_W[n]$ has dimensions $N_r \times N_c \times 3$.  The RGB output after applying this weakly phasic temporal filter within each channel is used as input to the color subchannels.  The color channel is made up of four subchannels.  These are red-green opponency ($RG$), green-red opponency ($GR$, blue-yellow opponency ($BY$), and yellow-blue opponency ($YB$).  These features are extracted by decoupling hue from intensity by normalizing each color channel by intensity.  These subchannels are computed using the temporal output as follows:

\begin{equation}
\label{eq:colorchannel_R}
R = \floor{r-\frac{g+b}{2}}
\end{equation}
\begin{equation}
\label{eq:colorchannel_G}
G = \floor{g-\frac{r+b}{2}}
\end{equation}
\begin{equation}
\label{eq:colorchannel_B}
B = \floor{b-\frac{r+g}{2}}
\end{equation}
\begin{equation}
\label{eq:colorchannel_Y}
Y = \floor{\frac{r+g}{2}-\frac{|r-g|}{2}-b}
\end{equation}

Using Equations \ref{eq:colorchannel_R} to \ref{eq:colorchannel_Y}, the four color opponencies are computed as follows:

\begin{equation}
RG = \floor{R-G}
\end{equation}
\begin{equation}
GR = \floor{G-R}
\end{equation}
\begin{equation}
BY = \floor{B-Y}
\end{equation}
\begin{equation}
YB = \floor{Y-B}
\end{equation}

These four separable spatio-temporal filtered outputs ($RG$,$GR$,$BY$,$YB$) are used as input to the grouping stage of the color channel.

\subsubsection{Orientation Channel - Spatial Content Only}
Within the orientation channel, there is no temporal filtering.  Extraction of temporal information within the intensity channel and color channel is sufficient.  Furthermore, this helps to preserve static information with regards to saliency.  In the orientation channel, there are four subchannels.  Within each channel, saliency is computed in regards to salient objects with respect to a unique orientation.   These four subchannels are $O_{0}$, $O_{\frac{\pi}{4}}$, $O_{\frac{\pi}{2}}$, and $O_{\frac{3\pi}{4}}$ where $0$, $\frac{\pi}{4}$, $\frac{\pi}{2}$, and $\frac{3\pi}{4}$ correspond to the four unique orientations.  For each of these subchannels, the grayscaled, intensity version of the current frame (see Equation \ref{eq:intensityextraction}) is the input to the grouping stage.

\subsection{Grouping Mechanism and Normalization}
The spatio-temporal output of each of these feature channels is fed as input to the grouping stage of the model.  The grouping mechanism is that used in \cite{russell2014model}.  This model is inspired by Craft et al. \cite{craft2007neural}.  The first stage of this mechanism is extraction of object edges, similarly to the receptive field of simple cells in V1.  Both odd and even responses are combined to form complex cell responses.  These complex cells are contrast-invariant edge responses which directly excite left or right side preferred border ownership neurons.  In order to extract information regarding the existence of objects, a center surround operation is performed.  This is similar to the receptive field of neurons found within the retina and LGN -- both ON- and OFF-center receptive fields.  This is necessary for detecting dark objects on light backgrounds as well as light objects on dark backgrounds.  The border ownership responses to the complex cells are modulated by the center-surround cell responses.  Excitation from the center-surround response coding for figure on the border ownership cells preferred side increases border ownership activity.  Center-surround activity on the non-preferred side inhibits/suppresses border-ownership activity.  Finally, border ownership activity is integrated in an annular fashion to give grouping cell activity.  This grouping activity is representative of dynamic proto-object activity giving rise to figure-ground relationship of the dynamic visual scene.

The same normalization techniques discussed earlier (seen in Itti et al. \cite{itti1998model}) are used across scale and across individual feature conscpicuity maps.  The results within each channel are linearly summed to form the instantaneous saliency map at a given time (i.e. frame).  These sequential saliency maps over time form the final dynamic saliency map.  This is seen in Fig.~\ref{fig:dpovsst}.

\section{FPGA Implementation of the Model}
\label{sec:podvs_model_fpga}
The objective of this hardware, parallel architecture implementation
is for real-time processing of the dynamic saliency map so it can
be utilized for real-world applications under low SWaP constraints.
To accomplish this, some components of the model did not require a
hardware implementation for real-time processing.  Therefore, we
utilize a hybrid approach in which some components of the model were
processed in software on PC (MATLAB), while more computationally-heavy
tasks were implemented on dedicated hardware (FPGA) in order to take advantage of pipelining and parallel processing.  A high-level processing flow of this hybrid approach is visualized in Fig.~\ref{fig:fpga_model_full}.

\subsection{Model Rescaling for the FPGA}
Considering the complexity of the PODVS model discussed in Section
\ref{sec:podvs_model} and limited resources on the Opal Kelly 7350
Kintex-7 FPGA, the model was rescaled for FPGA implementation.  The
processing steps of the model remain the same and the
biological-plausibility and basis of the model were not compromised.
The rescaled parameters for the FPGA implementation are outlined in
Table~\ref{tab:specs_matlabvsfpga}. 

\begin{table}[!htbp]
\caption{Summary of PODVS Model Rescaling for FPGA Implementation}
\label{tab:specs_matlabvsfpga}
\centering
\begin{tabular}{|c||c|c|}
\hline
\textbf{Model Specification} & \textbf{MATLAB Design} & \textbf{FPGA Design} \\
\hline
\hline
Kernel Size (Grouping) 		&  $11 \times 11$		&  $5 \times 5$	\\
\hline
\# Pyramid Levels 			&  $10$ 			&  $3$			    \\
\hline
Resolution (W $\times$ L)	&  $640 \times 480$ &  $112 \times 84$ (or $80 \times 60$)  \\
\hline
Feature Channels 			&  $9$ 				& $9$				\\
\hline
\end{tabular}
\end{table}

The most significant rescaling was the input video resolution.
The software implementation expects an input resolution of $640 \times
480$ pixels.  The FPGA implementation has a resolution of $112 \times
84$, as well as the option for $80 \times 60$ resolution.  The reduced
resolution is necessary because the FPGA has a limited amount of
onboard BRAM (Block Random Access Memory).  Furthermore, with lower
resolution, the number of pyramid levels was also reduced.  The reason
for computing on an image pyramid is to allow for scale-invariance.
The MATLAB implementation has a resolution of $640 \times 480$ at the
top level of the pyramid and scales down 9 levels.   Operating at a
resolution of $112 \times 84$ or $80 \times 60$ requires only 3
pyramid levels.  With reduced resolution, the kernel used in the
various filtering tasks within the grouping computation stage was also
rescaled.  In the software implementation, the kernel size is $11
\times 11$ for performing the various edge, center-surround, and von
Mises filtering, and related tasks.  Note, the von Mises filtering
step uses a kernel (generated using the von Mises distribution) to map
center-surround responses to edges for computing border-ownership activity.
This is explained in more detail in \cite{russell2014model}. In the FPGA
implementation, the filter size is $5 \times 5$ for these tasks within
the grouping mechanism stage.  This rescaling was necessary to
compensate for the lower resolution video frames.  

\subsection{FPGA Processing Blocks}
\begin{figure*}[!htbp]
\centering
\includegraphics[width=1.0\textwidth]{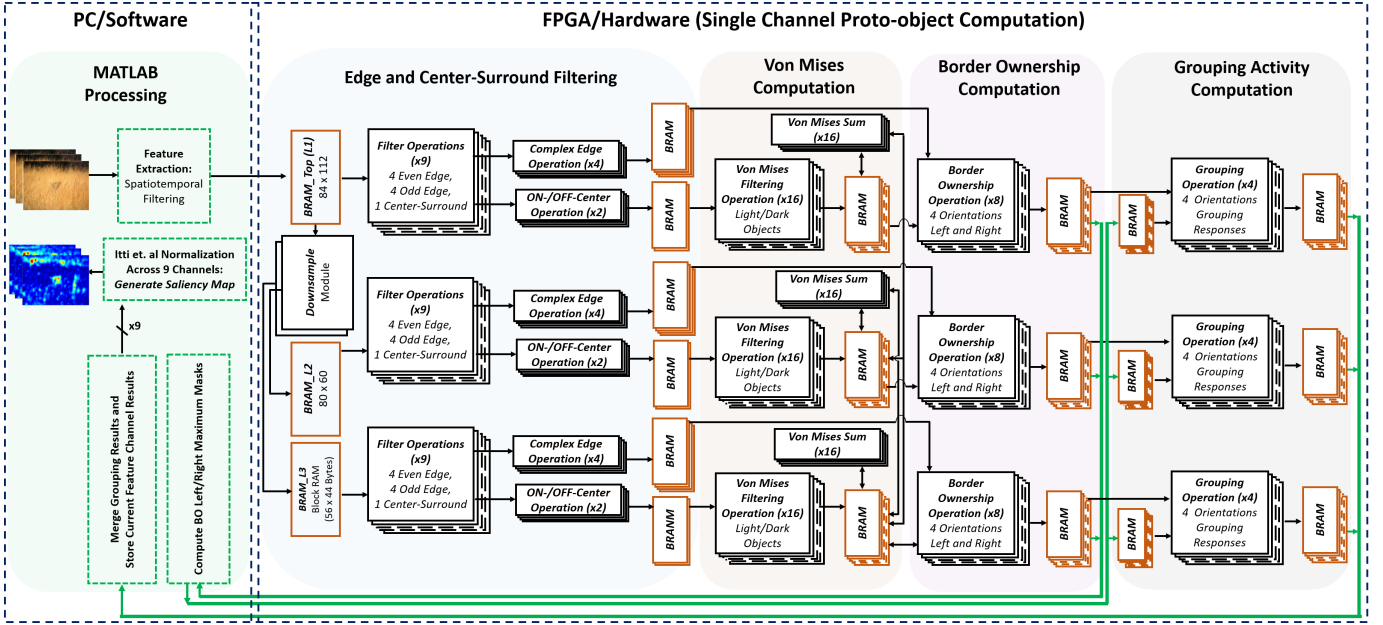}
\caption{Block diagram of the complete FPGA implementation of our Proto-object Based Dynamic Visual Saliency Model.}
\label{fig:fpga_model_full}
\end{figure*}

The detailed FPGA implementation can be visualized in
Fig.~\ref{fig:fpga_model_full} and will be discussed in this section.
The key components of the model are as follows: the input video
stream, spatial and temporal component extraction, grouping
computation, and normalization operator and output of dynamic saliency
map.  Grouping features into proto-objects requires some prior computations.  This includes the edge and
center-surround operations, von Mises computation, and border ownership computation.

\subsubsection{Input Video Stream}
The input to the model is a preloaded or live-stream video.  The
resolution expected by the model can be either $112 \times 84$ or $80
\times 60$ pixels.  The pixels are at 8-bit resolution.

\subsubsection{Feature and Motion Extraction}
Prior to transmitting frames to the FPGA, the spatial features for each of the 9 channels are extracted.  Within the intensity channel, the temporal information is extracted using strongly phasic receptive fields.  Within the color channel, the same 4 subchannels are extracted: red-green opponency, green-red opponency, blue-yellow opponency, and yellow-blue opponency.  Prior to extracting these color opponencies, motion is extracted using the weakly phasic temporal filter.    The spatial color opponencies are then extracted from this temporal output.  Similarly to the software implementation, motion is not extracted within the orientation channel.  The gray-scaled version of the current frame serves as input to the 4 orientation subchannels.

\subsubsection{Border Ownership and Grouping Computation}
The border ownership and grouping mechanism is
computationally-demanding, and therefore, processing of the grouping
activity predominantly occurs on the FPGA.  Grouping is
computed within each of the nine feature channels independently, 
making parallel processing of each channel on the FPGA advantageous.  In this section, we describe the method for computing the grouping activity within a single channel.  Identical processing occurs in parallel for each channel.  These steps are outlined below.

\begin{enumerate}[a)]
\item P1: Transmit Spatio-temporal Feature Extraction Output
\item P2: Generate Frame Pyramid
\item P3: Complex Edge and Center-Surround Filtering
\item P4: von Mises Filtering
\item P5: von Mises Sum
\item P6: Border Ownership Responses
\item P7: Grouping Responses
\end{enumerate}

\paragraph{P1}
This first step involves transferring the output of the
spatio-temporal feature extraction to the FPGA for processing.  This
is a single spatio-temporal response at the highest resolution
($112\times84$) with 8-bit precision per pixel.  This data is
transferred to BRAM on the FPGA.  The amount of BRAM required for
storing the response of each channel is $\sim75KB$ (see Equation \ref{eq:compute_transfer_bram}).
The USB 3.0 communication from the PC
to the Opal Kelly FPGA allows for a transmission speed of $340$ MB/s. 
\begin{equation}
\label{eq:compute_transfer_bram}
\text{Transmitted Data} = 112 \times 84 \times \texttt{8-bits} \approx 75.2 \text{ KB}
\end{equation}

\paragraph{P2}
The frame/image pyramid is generated using a "nearest-neighbor" downsampling method.  Computing the address of the pixel in the $112\times84$ frame from which to subsample is approximated using bit-shifting for multiplications and divisions.  This computed address is used to read the pixel value which is stored in additional BRAM for the two additional images in the pyramid at resolutions $80\times60$ and $56\times44$.  Each address computation, read from BRAM, and write to new BRAM takes 5 clock cycles (CC) per pixel.  Downsampling occurs in parallel for each additional image in the pyramid.  Therefore, the bottleneck is the time to downsample the larger of the two images ($80\times60$).  The total number of clock cycles required for this step is therefore 24000 CC ($240\mu s$ for a $100$MHz clock).  This can be seen in Equation \ref{eq:downsample_cc}.  The total additional BRAM required is $\sim7.2 \text{ KB}$.
\begin{equation}
\label{eq:downsample_cc}
\text{Downsampling CC} = 80 \times 60 \times 5 \text{ CC} = 24\text{K CC}
\end{equation}

\paragraph{P3}
The next stage in computation involves performing parallel filtering tasks with supporting computation for extracting the complex edge response and both ON-center-surround and OFF-center-surround responses within the current feature channel.  The complex edge responses are extracted by computing the square root of the even and odd edge responses.  Note that an IP core for this FPGA is used for computing the square root.  The parameters of these kernels are selected such that there are four $5 \times 5$ even edge kernels and four $5 \times 5$ odd edge kernels.  The parameters are also selected such that a single ON-center-surround kernel is used for both the Orientation channels and non-Orientation channels and the OFF-center-surround response is simply the inverted response of the ON-center-surround response.  For computing these complex edge and center-surround responses, an FSM (finite state machine) is used, which performs the following steps:

\begin{enumerate}
\item Load $5 \times 5$ patch at the current pixel location
\item Compute weighted-sum for 8 edge and ON-center-surround response
\item Compute square root for complex edge response of all 4 orientations
\item Invert ON-center-surround response for OFF-center-surround response
\item Resulting 6 responses are stored in BRAM
\end{enumerate}

The most computationally-demanding step of these is the weighted sum.   The weighted sum mathematically can be expressed

\begin{equation}
\label{eq:compute_weightedsum}
\text{Weighted Sum} = \sum_{r=1}^5 \sum_{c=1}^5 K(r,c) \times P(r,c)
\end{equation}

The values $r$ and $c$ are the row and column, respectively, of the $5 \times 5$ kernel ($K$) and $5 \time 5$ image patch ($P$).  This is implemented in hardware as a series of multiply-accumulates (MAC).  The weighted-sum is essentially a sequences of MACs to obtain the final result at each pixel location.  For each of the 4 even edge kernels, 4 odd edge kernels, and ON-center-surround kernel, the weighted-sums are computed in parallel.  Each of these 9 weighted-sum operations occurring in parallel consists of a single MAC module each.  The number of CC required for performing a single MAC operation is 3 CC.  Furthermore, it requires 25 MAC operations (75 CC) to compute the weighted-sum results for all 9 kernels (See Equation \ref{eq:mac}).

\begin{equation}
\label{eq:mac}
\text{Weighted Sum CC} = 3 \text{ CC} \times 5 \times 5 = 75 \text{ CC}
\end{equation}

These filter responses are stored in BRAM for each frame in the
pyramid, requiring $\sim100.3 \text{ KB}$ and $\sim1.9 \text{M CC}$.

\paragraph{P4}
The next stage of processing uses the output from the center-surround responses for both ON-center-surround (light objects on dark background) and OFF-center-surround (dark objects on light background) responses (for each pyramid level) to compute the von Mises filter responses (weighted sums).  The von Mises response is necessary for mapping edges to their corresponding center-surround responses.  The von Mises responses are then used for computing the border ownership responses for both left and right sides of the oriented border.  Utilizing edge responses for 4 different orientations, on 2 sides of each center-surround response (left and right), on 2 center surround responses (ON- and OFF-), equates to 16 different filtering operations for each pyramid level.  All of these filtering tasks occur in parallel and results are stored in 48 independent BRAM modules (16 for each pyramid level).  The total amount of BRAM required to store the responses is $\sim266.7 \text{ KB}$ and the total number of clock cycles required is $\sim1.9 \text{M CC}$.

\paragraph{P5}
The next stage of processing involves the von Mises summing across pyramid levels.  There exist 16 FSM running in parallel (8 for von Mises responses for light objects and 8 for von Mises responses for dark objects from previous section).  For each of these parallel 16 processes, the following steps occur:

\begin{enumerate}
\item Apply scaling factor to obtain pixel value in other pyramid level (3 CC)
\item Retrieve pixel value in bottom adjacent level (1 CC)
\item Multiply by factor $2^{-j}$ where $j$ is the pyramid level (1 CC)
\item Accumulate result for current pyramid level (1 CC)
\item Repeat prior steps for each lower pyramid level, $k$ such that $\rightarrow k \leq j$
\item Repeat all previous steps for each pyramid level $j$
\item Store results in BRAM used for von Mises filtering (1 CC)
\end{enumerate}

This process must occur for each pixel location in each image.  Note that no additional block RAM is required as the results of the von Mises sum replace the values in the von Mises filtering responses as they are no longer required for the remainder of the model.  The time required for computing the von Mises sum for each of the 16 parallel summations is $\sim241 \text{K CC}$.

\paragraph{P6}
The next stage in processing is computation of the border ownership responses for both light objects and dark objects (ON- and OFF-center responses, respectively) for both left and right-side border ownership ($\theta$ and $\theta + \pi$).    The left and right border ownership responses for 4 oriented edges/borders ($\theta$) are computed using Equations \ref{eq:fpga_boleftfin} and \ref{eq:fpga_borightfin}.  They are computed by summing the border ownership response for light objects and dark objects for each oriented edge independently.  This gives rise to polarity-invariance with respect to center-surround response.

\begin{equation}
\label{eq:fpga_boleftfin}
B^k_{Left}[\theta] = B^k_{Light,Left}[\theta] + B^k_{Dark,Left}[\theta]
\end{equation}

\begin{equation}
\label{eq:fpga_borightfin}
B^k_{Right}[\theta] = B^k_{Light,Right}[\theta] + B^k_{Dark,Right}[\theta]
\end{equation}

The responses $B^k_{Light,Left}[\theta]$, $B^k_{Light,Right}[\theta]$, $B^k_{Dark,Left}[\theta]$, and $B^k_{Dark,Right}[\theta]$ are computed using the von Mises summation responses from step P5.  The variable $k$ corresponds to the pyramid level.  There are 8 FSM used for each pyramid level ($\times3$) for computing the two border ownership responses (for left border ownership and right border ownership) at each of the 4 edge orientations, $\theta$ ( $0$, $\frac{\pi}{4}$, $\frac{\pi}{2}$, and $\frac{3\pi}{4}$), and each of the 3 pyramid levels, in parallel.  Therefore, $12$ left and right border ownership responses are computed in parallel.  This totals to 24 border ownership responses, and therefore, 24 additional BRAMs for storing the responses.  The total computational time for computing the left and right border ownership (4 edge orientations) for each pyramid level is $\sim56 \text{K CC}$ and requires $\sim133.3 \text{ KB}$ of BRAM.\\

\paragraph{P7}
To compute the final grouping activity, masks must be generated using a max operator on the border ownership activity that effectively determines the appropriate objects to which the edges (i.e. borders) belong to.  To compute these masks, the border ownership responses are transmitted back to the PC and the masks are computed on the PC and then transmitted to the FPGA.  Once these  binary masks are received by the FPGA ($BOMaskLeft^k_\theta(x,y)$ and $BOMaskRight^k_\theta(x,y)$) for all pyramid levels ($k$) and orientations ($\theta$), the grouping responses for left border ownership and right border ownership are computed independently using Equations \ref{eq:fpga_grpleft} and \ref{eq:fpga_grpright}, respectively.

\begin{equation}
\label{eq:fpga_grpleft}
\begin{split}
GrpLeft^k_\theta(x,y) = BOMaskLeft^k_\theta \otimes B^k_{Left}[\theta] * v_\theta \\
- w_p \times BOMaskLeft^k_\theta \otimes B^k_{Right}[\theta] * v_{\theta}
\end{split}
\end{equation}

\begin{equation}
\label{eq:fpga_grpright}
\begin{split}
GrpRight^k_\theta(x,y) = BOMaskRight^k_\theta \otimes B^k_{Right}[\theta] * v_{\theta + \pi}\\
- w_p \times BOMaskRight^k_\theta \otimes B^k_{Left}[\theta] * v_{\theta + \pi}
\end{split}
\end{equation}

The parameter $w_p$ is the weight of the inhibitory connection to the opposing side border ownership.  In this model, $w_p=1$.  Note that $v_{\theta}$ and $v_{\theta + \pi}$ are the von Mises summation responses for both left and right border ownership responses.  The operator $\otimes$ is an element-wise operation.  The final grouping step involves summing the left and right grouping responses as seen in Equation \ref{eq:fpga_grpsum}.

\begin{equation}
\label{eq:fpga_grpsum}
GrpSum^k_\theta(r,c) = GrpLeft^k_\theta(r,c) + GrpRight^k_\theta(r,c)
\end{equation}

This results in grouping responses ($GrpSum^k_\theta(r,c)$) for each orientation, $\theta = 0, \frac{\pi}{4}, \frac{\pi}{2}, and \frac{3\pi}{4}$, for each pyramid level, $k = 1,2,3,...,12$).  The total number of CC for the grouping operation, independent of the PC computation of the masks, is $1.9 \text{M CC}$ and total additional BRAM required to store the grouping activity is $\sim66.6 \text{ KB}$.

\subsubsection{Final Normalization and Saliency Map Generation}
The final normalization stage and dynamic saliency map generation is
identical to the final normalization operator used in \cite{russell2014model}.  This final Itti et
al. \cite{itti1998model} normalization and merging of pyramid levels
to generate the final saliency map is computed on PC as a function of
the grouping activity computed on the FPGA.  For the fastest
processing speed, computing grouping activity within all feature
channels in parallel is ideal.  The Opal Kelly FPGA used in this work
had sufficient resources for computing a single channel for
$112\times84$ video frames.  However, this exact processing was
computed for video frames of resolution $80 \times 60$ and this same
FPGA had sufficient resources for computing grouping activity on 3 out
of 9 channels in parallel.  Results with respect to total resources
and speed will be discussed in the proceeding sections. 

\section{Experimental Setup}
We confirmed validity of this work in two ways.  First, we validate
this novel dynamic visual saliency model (PODVS) by quantifying the
model's ability to predict human eye fixations on videos.  These
results are further compared with other SOTA
bottom-up, dynamic visual saliency models.  Secondly, we validate the
PODVS model FPGA implementation by quantifying the similarity of the
FPGA implementation's computed dynamic saliency to that of the
analogous MATLAB implementation. 

\subsection{Dataset}
\label{sec:dataset}
The dataset used for this work is the CRCNS (Collaborative Research in
Computational Neuroscience) dataset created by Itti et
al. \cite{itti_carmi}.  This dataset consisted of a total of 520 human
eye-tracking data traces of young adult human volunteers, both male
and female, watching complex video stimuli including TV programs,
outdoor videos, and video games.  This dataset consists of 8 different
subjects watching 50 different video clips ranging from 6 seconds to
90 seconds each (totaling $\sim$25 minutes of video).  The eye
tracker, ISCAN RK-464, recorded at a 240 Hz sampling rate with 9-point
calibration after every 5 clips.   Each video was played at a
framerate of $\sim$30 Hz.  The subjects' fixation locations 
were used to validate the saliency maps computed.  This dataset was
used both for validating the PODVS model's ability to predict human
eye fixations and compared with that of other SOTA models, as well as
to validate the similarity between the FPGA implementation of the
model and the analogous MATLAB model.  The smaller version of this
dataset, called ``MTV'', was not used.  More details on this dataset
can be found at \cite{itti_carmi}. 

\subsection{Metrics - Comparing with SOTA Models}
To quantify the ability of each saliency map to predict human eye fixations, we use two commonly used metrics for evaluating saliency maps.  The first is the area under the curve of the Receiver Operating Characteristic curve (AUC-ROC) \cite{green1966signal}.  The second is the Kullback Leibler Divergence (KLD) \cite{itti2005principled, itti2006bayesian}.  In their original state, these metrics are sensitive to edge effects due to the filtering operations of the algorithms \cite{zhang2009sunday}.  Therefore, these saliency models may introduce a center bias into the algorithm.  Furthermore, typically, the center of the video is naturally the viewer's focus of attention, even more so at the beginning of a video \cite{tatler2007central, vitu2004eye, parkhurst2003scene}.  To compensate for these center bias effects and provide a fair comparison between models, the metrics are modified to only use saliency values at human fixation points given that human's do not typically look near the edges of images.  By using only human fixation points, any center bias will affect the metrics of each model equally.  Additionally, for all models' saliency map output, each frame is normalized between 0 and 1.  These metric measurements are similar to the metrics used in \cite{russell2014model}, except they use dynamic video rather than static images.

\subsubsection{Area Under Curve - Receiver Operating Characteristic}
For the AUC-ROC evaluation metric, the saliency map is treated
as a binary classifier.  By varying the threshold and extracting true positives and
false positives at each threshold, an ROC curve can be constructed.
Similarly to that used by \cite{zhang2009sunday,russell2014model}, for
extracting true positives, ground truth eye fixation points for the
frame being evaluated are used.  For extracting false positives,
random eye fixation points from other videos in the dataset are used.
This process is repeated 100 times for each frame and the average
score is computed, and further, the average is computed across all
frames to determine the score for a given video.  The final AUC-ROC
score is the average of that over each video in the dataset.  A score of 1 represents perfect prediction, 0.5 represents chance, and less than 0.5 represents anti-correlation.  This score is then normalized by the AUC-ROC score describing the ability of human fixations to predict other human fixations \cite{russell2014model,masciocchi2009everyone}.

\subsubsection{Kullback Leibler Divergence}
The KLD evaluation metric is effectively the ``difference'' between two distributions.  The first is a histogram of the saliency values sampled at true fixation points of the frame being evaluated.  The second is a histogram of saliency values computed at random fixation points in the same frame.  This is modified (similarly to \cite{zhang2009sunday,russell2014model,tatler2007central}) such that the random fixation points are taken from randomly selected fixation points from other videos in the dataset.  A higher KLD value is better.  The same normalization method used in computing the the AUC-ROC score is used for normalizing the KLD score for each video and for each model \cite{masciocchi2009everyone}.

\subsection{Metrics - Validating the FPGA Implementation}
The purpose of the FPGA implementation of this model is to speedup processing while producing the same output as that of the software implementation.  In order to validate the similarity of the FPGA implementation's dynamic saliency map output to the software implementation's output, we use the Pearson correlation coefficient \cite{pearson1896vii} and normalized scanpath saliency \cite{peters2005components} metrics.

\subsubsection{Pearson Correlation Coefficient}
The Pearson's correlation coefficient (PCC) can be used to measure the
linear correlation between two distributions \cite{pearson1896vii}.
We use the PCC metric to evaluate the linear relationship between the FPGA implementation's saliency map and the software implementation's saliency map output on visual stimulus.

\subsubsection{Normalized Scanpath Saliency}
We use the normalized scanpath saliency (NSS) metric
\cite{peters2005components} to measure the saliency values of the FPGA
implementation's dynamic saliency map at the salient locations
computed by the software implementation's saliency map output.  Both
saliency maps are normalized between 0 and 1.  We then use the
software implementation's saliency map and threshold the map at a
value of 0.7, creating a binary fixation map.  This map is then used
to determine the saliency values of the FPGA implementation at these
fixation locations.  A value of 1 means the saliency map perfectly
predicts fixation locations predicted by the software implementation's
saliency map.  Therefore, an NSS score closer to 1 means the FPGA
implementation predicts saliency similarly to that of the software
implementation. 

\begin{figure*}[!htbp]
\centering
\includegraphics[width=.75\textwidth]{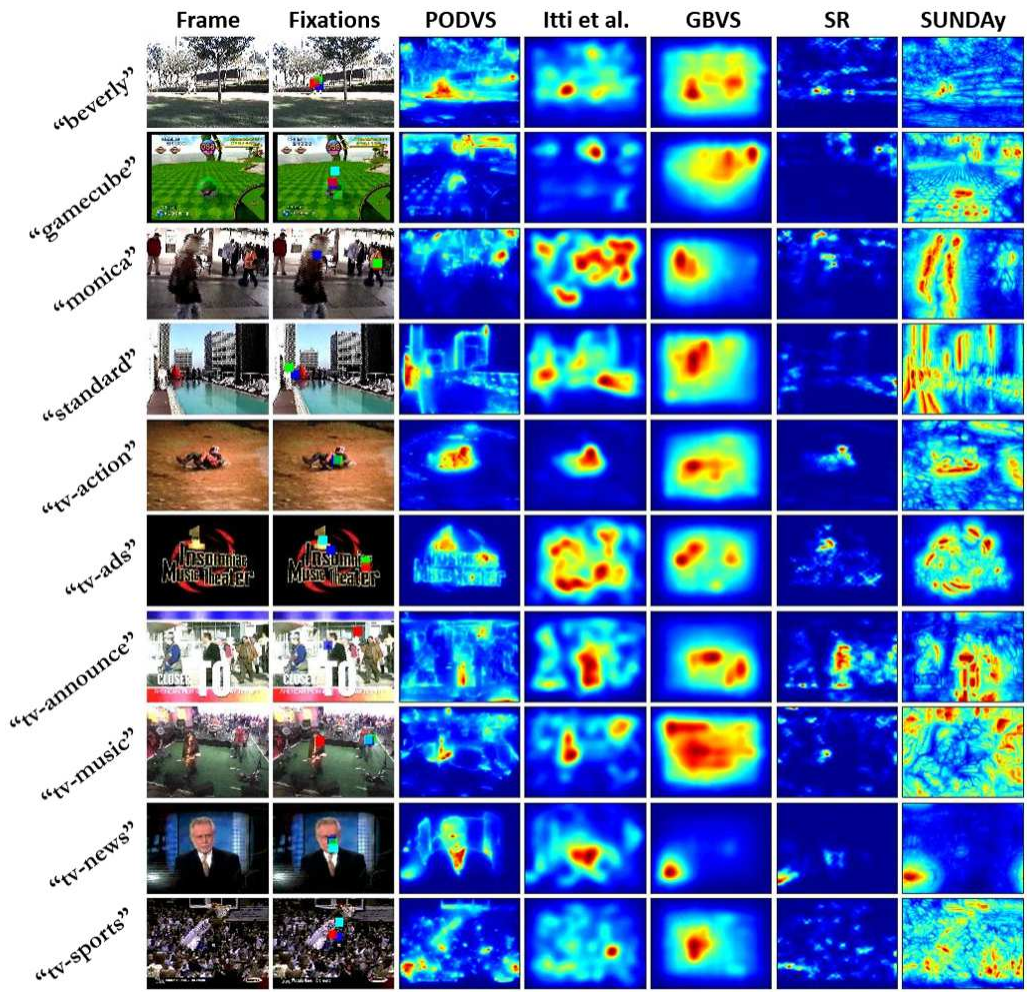}
\caption{Comparison of dynamic visual saliency map output with SOTA
  models.  Each row signifies a single frame ($480 \times 640$) from a
  single video from a unique class from the CRCNS dataset noted by the
  left/vertical labels.  The first column, ''Frame'', is the frame on
  which saliency was computed.  The second column, ''Fixations'', is
  the same frame overlaid with colored squares signifying the
  locations of subjects' fixations at that frame.  Different colors correspond to different observers.
  Columns three through seven are the dynamic saliency map outputs at the frame for
  our PODVS model and the Itti et al. \cite{itti2002real}, GBVS
  \cite{harel2007graph}, SR \cite{seo2009static}, and SUNDAy
  \cite{zhang2009sunday} models, respectively.} 
\label{fig:model_compare}
\end{figure*}

\section{Results and Discussion}
\subsection{Comparison With SOTA}
We first compare our model, PODVS, with four SOTA bottom-up, dynamic
visual saliency models.  Each of these models has
been previously
described.  The first is the Itti et al. \cite{itti2002real} model,
similar to \cite{itti1998model} but with
an additional motion channel.  The second is the
graph-based dynamic visual saliency model (GBVS) by Harel et
al. \cite{harel2007graph}, which is similar to the Itti et al. model
but reformulated with a graph-based approach.  A variant of this model was implemented in MATLAB which includes a motion-sensitive channel \cite{harel2006saliency}.
The third is the model by Seo and Milanfar
\cite{seo2009static} which uses a ``self-resemblance'' (SR) measure to
compute the likeliness of saliency at a each pixel.  The fourth model
is the Zhang et al. \cite{zhang2009sunday} model in which saliency is
computed using natural statistics for dynamic analysis of scenes
(SUNDAy).  As previously discussed, the metrics used for comparing our
model with these four SOTA models are AUC-ROC and KLD. 

A visual comparison of the saliency output of the different models can be seen in Fig.~\ref{fig:model_compare}.  Each column contains the heatmap representation of the saliency map output of a different model.  Red represents high saliency and blue represents low saliency.  For each model, the saliency map output is the output of a single frame of a randomly selected video from the dataset.  The CRCNS dataset contains various different classes of videos.  Each row are saliency map outputs of a frame from a video from a unique class of videos from the dataset: ''beverly``, ''gamecube``, ''monica``, ''standard``, ''tv-action``, ''tv-ads``, ''tv-announce``, ''tv-music``, ''tv-news``, and ''tv-sports``.  The first column, ''Frame``, displays the selected frame on which the dynamic saliency model was computed.  The ''Fixations`` column is the frame overlaid with subjects' eye fixations denoted by the colored squares.

Fig.~\ref{fig:model_compare} only reveals a qualitative comparison of
the output of the different models, and therefore, it is necessary to
do a quantitative comparison.  The comparison of the AUC-ROC and KLD
metrics can be seen in Table~\ref{tab:comp_to_sota}.  These results
are the mean AUC-ROC and mean KLD scores across all 50 videos in the
dataset.  Each row represents a different model and each column
represents a different metric.  The last column shows the p-values
after performing a t-test between the PODVS distribution of scores and
the corresponding model.  A p-value less than 0.01 ($10^{-2}$) means
that the comparison to our model is statistically significant.  For
the AUC-ROC score, all models performed better than chance.  It can be
seen that our model, PODVS, performed significantly better (with
respect to p-value) than all four other SOTA models for both the
AUC-ROC score ($AUC-ROC = 0.6745$) and KLD score ($KLD = 0.3507$).
The Itti et al. had slightly lower scores ($AUC-ROC = 0.6636$, $KLD =
0.3457$) than our PODVS model, but higher than the other three models.
The closeness in performance of our PODVS model and the Itti et al. model may be due to the same normalization operator used in both models.
However, our model computes the notion of dynamic proto-objects prior
to this normalization operator and performs better, hence, supporting
Gestalt psychology and the idea that attention is object-based, not
feature-based. 

\begin{table}[!htbp]
\caption{Average AUC-ROC and KLD Scores of PODVS and Other SOTA Models on CRCNS Dataset}
\label{tab:comp_to_sota}
\centering
\begin{tabular}{|l||c|c||c|c|}
\hline
\textbf{Model} 							& \textbf{AUC-ROC} 	& \textbf{p-value} & \textbf{KLD} & \textbf{p-value} \\
\hline
\hline
Chance  	& $0.5$  & $-$	&  $0$	 & $-$ \\
\hline
\textbf{PODVS} 		& \textbf{$\boldsymbol{0.6745}$} & $-$	&  \textbf{$\boldsymbol{0.3507}$} & $-$  \\
\hline
Itti et al. \cite{itti2002real}  	& $0.6636$ & $<10^{-12}$	&  $0.3457$ & $<10^{-8}$  \\
\hline
GBVS \cite{harel2006saliency}		& $0.5859$ & $<10^{-12}$	&  $0.3092$ & $<10^{-12}$  \\
\hline
SR \cite{seo2009static}		& $0.5595$ & $<10^{-12}$	&  $0.2522$ & $<10^{-12}$  \\
\hline
SUNDAy \cite{zhang2009sunday}		& $0.6080$ & $<10^{-12}$ 	&  $0.3025$ & $<10^{-12}$  \\
\hline
\end{tabular}
\end{table}

\subsection{Evaluation of the FPGA Implementation}
\subsubsection{Saliency Map Output Accuracy}
\begin{figure}[ht]
\centering
\includegraphics[width=0.45\textwidth]{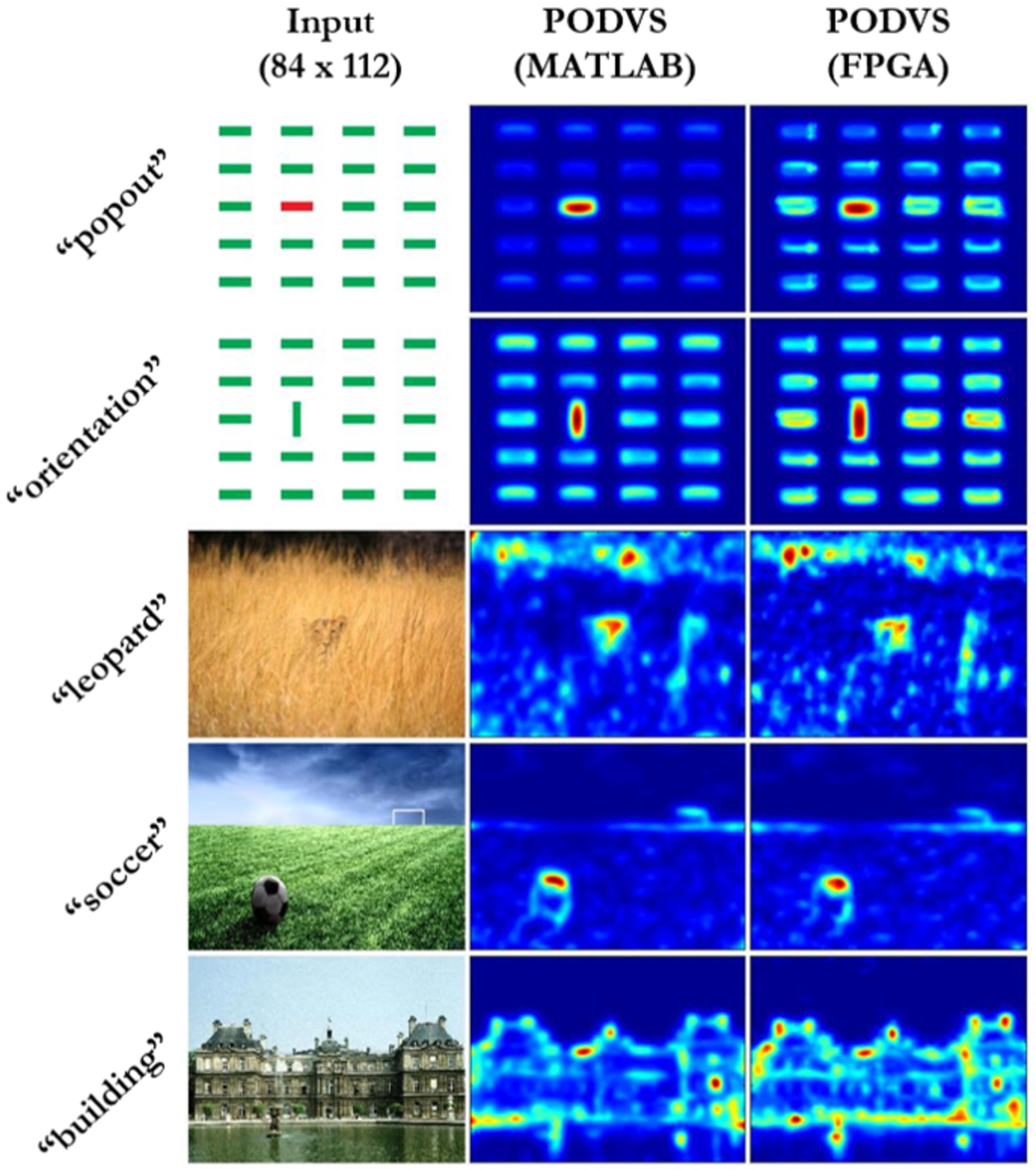}
\caption{Saliency map output of our PODVS model's FPGA implementation and MATLAB implementation (modified to match the specifications of the FPGA implementation) on $112 \times 84$ resolution input.  The first column is the input, the second column is the output of the MATLAB implementation.  The third column is the output of the FPGA implementation.}
\label{fig:fpga_matlab_compre}
\end{figure}

A visual comparison of the MATLAB implementation's saliency map output
to the FPGA implementation's output can be seen in
Fig.~\ref{fig:fpga_matlab_compre}.  For clear visual comparison, multiple static images were used as input to the MATLAB implementation and FPGA implementation.
The FPGA implementation of the PODVS model contains modified
parameters due to the limitations of the Opal Kelly Kintex-7 FPGA.
For fair comparison, the same modifications made to the FPGA
implementation were also made to the MATLAB implementation.  Ideally,
the output of the MATLAB implementation compared to the FPGA
implementation should be identical, however, the MATLAB implementation
uses floating-point precision for its computations while the FPGA uses
fixed-point precision for its computation due to the digital nature of
the FPGA hardware.  As seen in Fig.~\ref{fig:fpga_matlab_compre}, for each input image, 
the output of the MATLAB and FPGA implementation's are similar.  Although visually they are similar, we quantify the similarity for verification.

To quantify the actual similarity in saliency model output between the
FPGA implementation and MATLAB implementation, the PCC and NSS metrics
are used.  The PCC metric measures the similarity between the MATLAB
version's output and the FPGA version's output.  A PCC value of 1
means that there is perfect correlation, i.e. the two saliency maps
are identical.  A PCC value of 0 means no correlation.  Therefore, the
closer the PCC score is to 1, the better.  However, as noted, there
will be degradation in the precision of the FPGA implementation's
dynamic saliency map output due to its fixed-point precision.  The PCC
score of the FPGA implementation's output with respect to the MATLAB
implementation's output can be seen in
Table~\ref{tab:comp_fpga_matlab}.  To compute the PCC and NSS scores, both the $112\times84$ resolution FPGA implementation and $80\times60$ resolution implementation were ran on the same CRCNS dataset described in Section \ref{sec:dataset}.  The average PCC and NSS scores were then computed across all videos for each.  Table~\ref{tab:comp_fpga_matlab}
shows the average PCC score for both the $112\times84$ frame resolution input
and $80\times60$ frame resolution.  The average PCC score for the
$112\times84$ resolution version is $0.83$, and the average PCC
score for the $80\times60$ resolution version is $0.80$.
Furthermore, the NSS score quantifies the similarity in how the FPGA
implementation and MATLAB implementation's saliency map outputs
predict fixation points.  As can be seen
in Table~\ref{tab:comp_fpga_matlab}, for both resolutions the NSS scores
are close to 1.  This further confirms that the FPGA implementation
is close to that of the MATLAB implementation.  Also, the $80\times60$ resolution version has a lower average PCC and NSS scores than the $112\times84$ resolution version due to the reduced image quality at lower resolutions affecting the FPGA's ability to properly compute grouping activity using fixed-point precision versus that of MATLAB's floating-point precision.

\begin{table}[!htbp]
\caption{Average PCC and NSS Scores Quantifying the Similarity between the FPGA and MATLAB Implementation}
\label{tab:comp_fpga_matlab}
\centering
\begin{tabular}{|l||c|c|}
\hline
\textbf{Implementation} & \textbf{Average PCC Score}	& \textbf{Average NSS Score}\\
\hline
  \hline
\textbf{FPGA - ($\bf{112\times84}$)} 		& $0.83$ & $0.92$  \\
\hline
\textbf{FPGA - ($\bf{80\times60}$)} 		& $0.80$ & $0.90$  \\
\hline
\end{tabular}
\end{table}

\subsubsection{Resources}
The resources used for this complete model for both the version for $112 \times 84$ and $ 80 \times 60$ resolution version of the model can be seen in Table~\ref{tab:fpga_vs_resources}.  The limiting resources were the available DSP (Digital Signal Processing) slices and BRAM.  Instantiating another module for computing the grouping activity of another feature channel in parallel at $112 \times 84$ resolution would utilize too many resources and would not fit on this FPGA.  However, for the $80 \times 60$, three parallel feature channels is sufficient for fitting on this FPGA.

\begin{table}[!htbp]
\caption{Resources Used by OK XEM7350-160T FPGA for PODVS}
\label{tab:fpga_vs_resources}
\centering
\begin{tabular}{|c||c|c|c|}
\hline
\textbf{Resource} & \textbf{Available} & \textbf{Used ($112 \times 84$}) & \textbf{Used ($80 \times 60$)} \\
\hline
\hline
\textbf{Slice Registers}  					& 202,800 	& 33,911 (16\%)	& 17,764 (8\%)\\
\hline
\textbf{Slice LUTs}  						& 101,400 	& 36,313 (35\%) & 20,166 (19\%)\\
\hline
\textbf{BRAM (B36E1)}  				& 325 		& 195 (60\%)    & 139 (42\%)\\
\hline
\textbf{BRAM (B18E1)}  				& 650 		& 179 (27\%) 	& 180 (27\%)	\\
\hline
\textbf{DSP48E1 Slices}  				& 600 		& 594 (99\%) 	& 339 (56\%)	\\
\hline

\end{tabular}
\end{table}

\subsubsection{Speed}
At a resolution of $112 \times 84$, the FPGA-based PODVS model has a
framerate of $2.079 Hz$ for this Opal Kelly FPGA running on a $100$MHz
clock. The MATLAB version of this model (also with equally rescaled
parameters) has a runtime of $\sim1.27s$ on an Intel Quad-Core i7 PC.
This is a $2.64\times$ speedup in computation time.  This is a
significant speedup considering it only processes a single channel in
parallel on the FPGA.  Given an FPGA with sufficient resources for
computing all 9 channels in parallel, this FPGA implementation has a
$23.77\times$ speedup (at a framerate of $18.71$Hz) with respect to
the corresponding MATLAB implementation.  For $80\times60$ resolution
video frames, this Opal Kelly FPGA has enough resources to process 2
channels in parallel.  Therefore, at this resolution on the Opal Kelly
FPGA, it can compute saliency maps at framerate of $5.19 Hz$, which is
a $5.87\times$ speedup with respect to its corresponding MATLAB
implementation, also equally rescaled and at the same resolution.
Given an FPGA with sufficient resources to process all 9 channels in
parallel for $80\times60$ resolution video frames, this FPGA
implementation can achieve a $26.41\times$ speedup (at a framerate of
$23.35$Hz) compared to its corresponding MATLAB implementation
framerate.  To calculate the amount of sufficient resources, it is
assumed that the resources scales roughly linearly with the
number of channels being processed in parallel.  Therefore, for the
$112\times84$ resolution version, the resource utilization noted in
Table~\ref{tab:fpga_vs_resources} can be multiplied by 9 for
estimating the total amount of resources required for parallel
processing of all 9 channels.  Similarly, for the $80\times60$
resolution version, the resource utilization noted in
Table~\ref{tab:fpga_vs_resources} can be multiplied by a factor of 4.5
to determine the amount of resources required to process all 9
channels in parallel.  An FPGA such as a Virtex 7 would be sufficient
for processing all 9 channels in parallel at either resolution.  This
significant speedup in dynamic saliency map computation and
small-size of the FPGA, makes this FPGA implementation suitable for real-world
applications requiring real-time processing.  Table
\ref{tab:fpga_vs_speed} summarizes these results. 
\begin{table}[!htbp]
\caption{Speed Comparison of FPGA Implementation With MATLAB Implementation}
\label{tab:fpga_vs_speed}
\centering
\begin{tabular}{|p{4cm}||c|c|}
\hline
\textbf{Model} 							& \textbf{Framerate} 	& \textbf{Speedup} \\
\hline
\hline
\textbf{OK Kintex-7 FPGA - $112\times84$}  	& $2.079$ Hz 	&  $2.64\times$	 \\
\hline
\textbf{Ideal FPGA$\text{*}$ - $112\times842$}  		& $18.71$ Hz 	&  $23.77\times$	 \\
\hline
\textbf{OK Kintex-7 FPGA - $80\times60$}  	& $5.190$ Hz 	&  $5.87\times$ \\
\hline
\textbf{Ideal FPGA$\text{*}$ - $80\times60$}  		& $23.35$ Hz 	&  $26.41\times$	 \\
\hline
\end{tabular}
\vspace{0.05cm}
\\ \scriptsize $\text{ *}$ = FPGA with sufficient resources to process 9 channels in parallel.
\end{table}
\section{Conclusion}
In conclusion, we report on two advances.
We first present a novel dynamic visual saliency model, PODVS, based on the notion of dynamic
proto-objects that exist preattentively within the scene.  This neuromorphic model
is feed-forward, bottom-up, and biologically-plausible in its
computation, suggesting how dynamic visual saliency is computed in the early stages of
human visual processing.
Our neuromorphic model outperforms other SOTA dynamic visual saliency models in predicting
human fixations on videos and no training on large datasets is required.  Secondly, we present a novel FPGA
implementation for real-time, low power processing of this PODVS model
to be used for real-world applications.  The FPGA implementation
allows for up to $\sim26\times$ speedup compared to that of its
analogous CPU implementation, while maintaining high similarity
in its output with respect to the MATLAB implementation.  This
work may serve as the foundation for future work which incorporates dynamic proto-objects computation in the presence of
dynamic visual stimuli for higher-level, top-down tasks such as image
detection, tracking, and classification in which learning is involved.
Furthermore, the biofidelic nature of this work makes this
model suitable for processing on neuromorphic hardware in a
spike-based manner to further achieve the low SWaP specifications we
seek \cite{thakur2018large,thakur2017neuromorphic}.

\section*{Acknowledgment}
This work was funded and supported by the Office of Naval Research under MURI (Multidisciplinary University Research Initiative) Grant N000141010278, the National Institutes of Health Grant R01EY027544, and the SERB (Science and Engineering Research Board) Grant, India: ECR/2017/002517.

\ifCLASSOPTIONcaptionsoff
  \newpage
\fi

\bibliographystyle{IEEEtran}
\bibliography{references}

\begin{IEEEbiography}[{\includegraphics[width=1in,height=1.25in,clip,keepaspectratio]{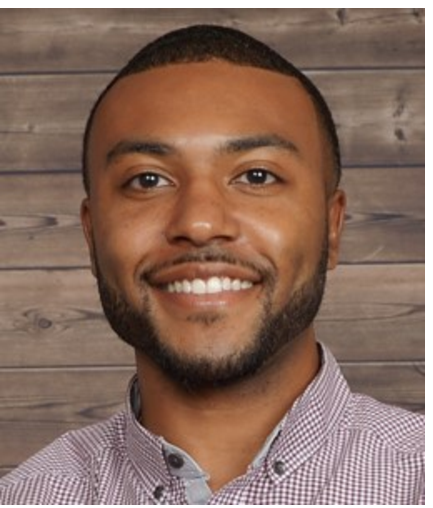}}]{Jamal Lottier Molin}
Dr. Jamal Lottier Molin received the B.S. degree in computer engineering from University of Maryland, Baltimore County (UMBC), Baltimore, MD, USA, in 2011, and the M.S.E and Ph.D. degrees in electrical and computer engineering from the Johns Hopkins University, Baltimore, MD, USA, in 2015 and 2017.  At UMBC, he was a recipient of the Meyerhoff Scholarship.  He also received the Department of Defense SMART (Science Mathematics and Research for Transformation) fellowship.  He serves as a review editor for Frontiers of Neuroscience in Neuromorphic Engineering.  He currently works for Riverside Research doing research in the field of neuromorphic engineering for vision and learning applications.\end{IEEEbiography}

\begin{IEEEbiography}[{\includegraphics[width=1in,height=1.25in,clip,keepaspectratio]{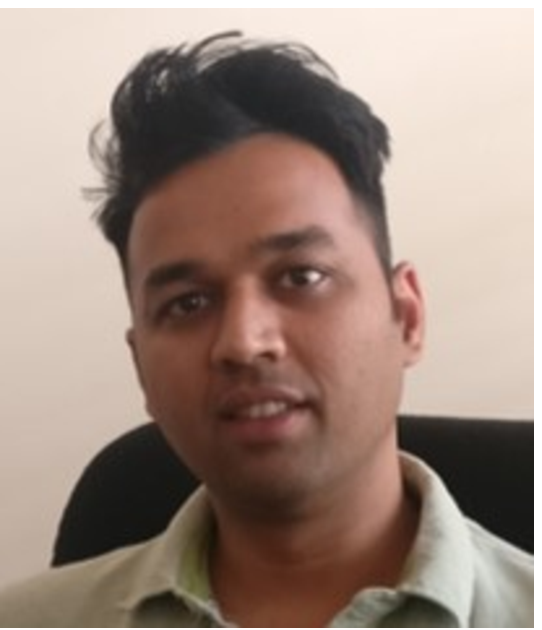}}]{Chetan Singh Thakur}
Dr. Chetan Singh Thakur is an Assistant Professor at Indian Institute of Science (IISc), Bangalore, and has an Adjunct Faculty appointment at the International Center for Neuromorphic Systems, Western Sydney University (WSU), Australia. Dr. Chetan Singh Thakur (Senior Member, IEEE) received his Ph.D. in neuromorphic engineering at the MARCS Research Institute, Western Sydney University in 2016. He then worked as a research fellow at the Johns Hopkins University. In addition, Dr. Thakur has extensive industrial experience. He worked for 6 years with Texas Instruments Singapore as a senior Integrated Circuit Design Engineer, designing IPs for mobile processors. His research expertise lies in neuromorphic computing, mixed-signal VLSI systems, computational neuroscience, probabilistic signal processing, and machine learning. His
research interest is to understand the signal processing aspects of the brain and apply those to build novel intelligent systems. He is recipients of several awards such as Young Investigator Award from Pratiksha Trust, Early Career Research Award by Science and Engineering Research Board- India, Inspire Faculty Award by Department of Science and Technology- India.\end{IEEEbiography}

\begin{IEEEbiography}[{\includegraphics[width=1in,height=1.25in,clip,keepaspectratio]{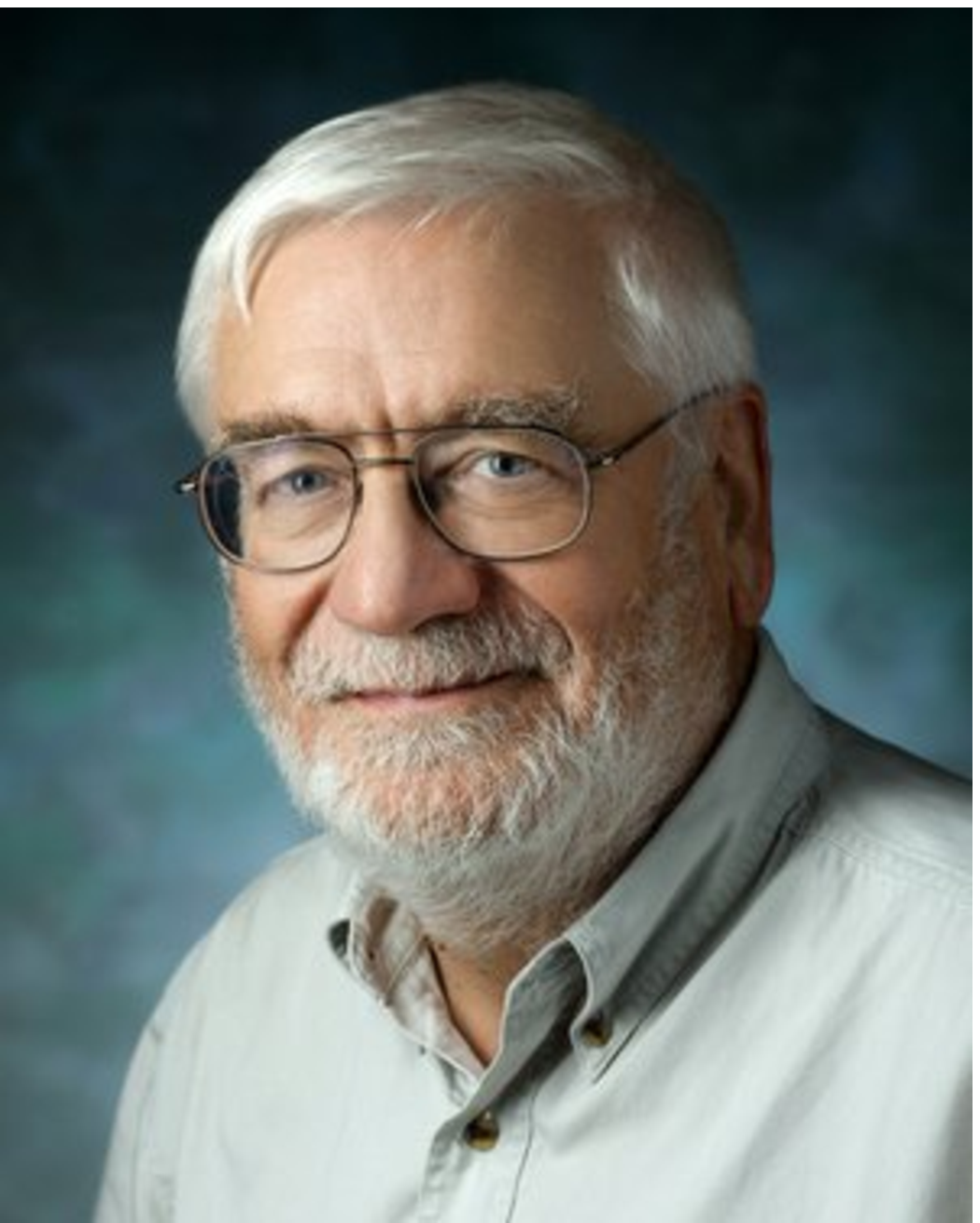}}]{Ernst Niebur}
  Dr. Ernst Niebur received the M.S. degree (Diplom Physiker) from the Universit\"{a}t Dortmund, Dortmund, Germany, the Postgraduate Diploma in artificial intelligence from the Swiss Federal Institute of Technology (EPFL), Lausanne, Switzerland, and the Ph.D. degree (Dr \`{e}s sciences) in physics from the Universit\'{e} de Lausanne, Lausanne, Switzerland.  His dissertation topic was a detailed computational model of the motor nervous system of the nematode C. elegans.  He was a Research Fellow and a Senior Research Fellow at the California Institute of Technology, Pasadena, CA, USA, and an Adjunct Professor at Queensland University of Technology, Brisbane, Australia.  He joined the faculty of Johns Hopkins University, Baltimore, MD, USA, in 1995, where he is currently a Professor of Neuroscience in the School of Medicine, and of Brain and Psychological Sciences in the School of Arts and Sciences. He is also Affiliated Faculty of the Institute for Computational Medicine, Whiting School of Engineering.  He uses computational neuroscience to understand the function of the nervous system at many levels.  Prof. Niebur is the recipient of a Seymour Cray (Switzerland) Award in Scientific Computation, an Alfred P. Sloan Fellowship, and a National Science Foundation CAREER Award.\end{IEEEbiography}


\begin{IEEEbiography}[{\includegraphics[width=1in,height=1.25in,clip,keepaspectratio]{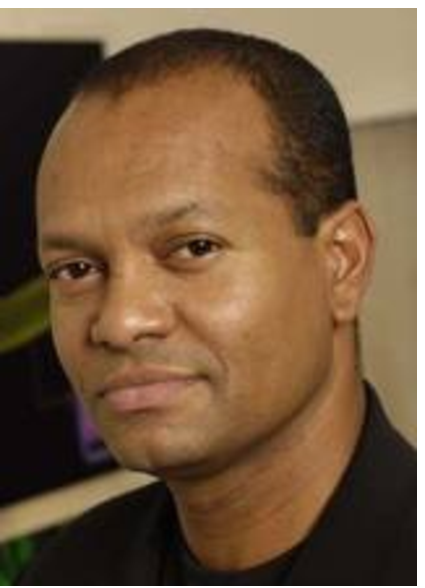}}]{Ralph Etienne-Cummings}
Dr. Ralph Etienne-Cummings (F'13) received the B.S. degree in physics from Lincoln University, Lincoln, PA, USA, in 1988, and the M.S.E.E. and Ph.D. degrees in electrical engineering from the University of Pennsylvania, Philadelphia, PA, USA, in 1991 and 1994, respectively. He is currently a Professor of electrical and computer engineering, and computer science with The Johns Hopkins University, Baltimore, MD, USA. He was the Founding Director of the Institute of Neuromorphic Engineering. He has authored more than 200 peer-reviewed articles and holds numerous patents. He has served as the Chairman of various IEEE Circuits and Systems (CAS) Technical Committees and was elected as a member of CAS Board of Governors. He also serves on numerous editorial boards. He is a recipient of the NSFs Career and Office of Naval Research Young Investigator Program Awards. He was a Visiting African Fellow at the University of Cape Town, Fulbright Fellowship Grantee, Eminent Visiting Scholar at the University of Western Sydney and has also received numerous publication awards, including the 2012 Most Outstanding Paper of the IEEE Transactions on Neural Systems and Rehabilitation Engineering. In addition, he was recently recognized as a Science Maker, an African American history archive.\end{IEEEbiography}




\end{document}